\documentclass{article} 
\usepackage{nips13submit_e,times}
\usepackage[utf8]{inputenc}
\usepackage{hyperref}
\usepackage{url}
\usepackage{graphicx}
\usepackage[numbers]{natbib}
\usepackage{indentfirst}
\usepackage{hhline}
\usepackage{tikz}
\usepackage{amsmath,amssymb}
\usepackage[toc,page]{appendix}

\title{CNN-based Segmentation of Medical Imaging Data}

\author{
Bar\i\c{s}~Kayal\i bay \thanks{Bar\i\c{s}~Kayal\i bay is affiliated with Faculty of Informatics, TUM}\\
\And 
Grady Jensen\thanks{Grady is affiliated with the Bernstein Center for Computational Neuroscience and the Graduate School of Systemic Neurosciences, Ludwig-Maximilians-Universit{\"a}t M{\"u}nchen, Germany} \\
\And 
Patrick van der Smagt\thanks{Patrick van der Smagt is affiliated with Data Lab, Volkswagen Group, M\"unchen, Germany.} \\
}

\nipsfinalcopy 

\begin{document}

\maketitle

\begin{abstract}
Convolutional neural networks have been applied to a wide variety of computer vision tasks. Recent advances in semantic segmentation have enabled their application to medical image segmentation. While most CNNs use two-dimensional kernels, recent CNN-based publications on medical image segmentation featured three-dimensional kernels, allowing full access to the three-dimensional structure of medical images. Though closely related to semantic segmentation, medical image segmentation includes specific challenges that need to be addressed, such as the scarcity of labelled data, the high class imbalance found in the ground truth and the high memory demand of three-dimensional images.

In this work, a CNN-based method with three-dimensional filters is demonstrated and applied to hand and brain MRI. Two modifications to an existing CNN architecture are discussed, along with methods on addressing the aforementioned challenges. While most of the existing literature on medical image segmentation focuses on soft tissue and the major organs, this work is validated on data both from the central nervous system as well as the bones of the hand.
\end{abstract}

\section{Introduction and Related Work}
\label{chapter:Intro}

Convolutional neural networks \cite{lenet} are a type of deep artificial neural networks widely used in the field of computer vision. They have been applied to many tasks, including image classification \cite{lenet,vggnet,alexnet,inception}, superresolution \cite{superres} and semantic segmentation \cite{fcn}. Recent publications report their usage in medical image segmentation and classification \cite{ciresan,unet,vnet,3dunet,kamnitsas,braint}. Medical images may come from a variety of imaging technologies such as computed tomography (CT), ultrasound, X-ray and magnetic resonance imaging (MRI). They are used in the depiction of different anatomical structures throughout the human body, including bones, blood vessels, vertebrae and major organs. As medical images depict healthy human anatomy along with different types of unhealthy structures (tumors, injuries, lesions, etc.), segmentation commonly has two goals: delineating different anatomical structures (such as different types of bones) and detecting unhealthy tissue (such as brain lesions). \par
This work demonstrates a CNN-based medical image segmentation method on hand and brain MRI, with the tasks of bone and tumor segmentation respectively. To this purpose, a network architecture similar to the ``U-Net'' architecture of Ronneberger et al.~\cite{unet} is used. Two modifications to the original design are tested:
\begin{enumerate}
\item Combining multiple segmentation maps created at different scales;
\item Using element-wise summation to forward feature maps from one stage of the network to the other.
\end{enumerate}
These modifications are discussed in Section~\ref{chapter:Method} and judged in terms of their performance in Section~\ref{chapter:Results}\footnote{We provide the code we used in our experiments on \url{https://github.com/BRML/CNNbasedMedicalSegmentation}}. The data used in the course of this work, as well as the scope and set-up of the experiments is explained in Section~\ref{chapter:Experiments}. This section continues with a survey of the recent advances in semantic segmentation before elaborating on the ``U-Net'' architecture, along with other approaches to CNN-based medical image segmentation.
\subsection{Background and Related Work}
\subsubsection{Advances in Image Segmentation}
Typical classifier deep neural networks like AlexNet \cite{alexnet}, VGGNet \cite{vggnet} or GoogLeNet \cite{inception} read in an image and output a set of class probabilities regarding the entire image. This is done by providing the raw image as input to the first layer of the network and performing a ``forward pass'': Sequentially letting each layer compute its respective function on the output of the layer that preceded it. An example for a classification problem is handwritten digit recognition, where a neural network receives an image of a single handwritten digit and must decide which of the ten digits is depicted on the image.  Segmentation tasks, on the other hand, require a classification at every pixel of the input. This goal can be met by applying a classifier to patches of the input in a sliding window fashion. In this scheme, each pixel is classified by extracting a patch around it and having it classified by the network. Multiple pixels can be classified at once by processing multiple overlapping patches as a batch and parallelizing the forward pass of each patch through the network. As nearby pixels in an image are similar in their vicinity, this approach suffers from redundant computation. Another limitation of this method is that the learning process is restricted to features that are visible in each patch and the network has less of a chance to learn global features the smaller the patch size is. This is supported by reports of larger patch sizes yielding better results \cite{ciresan,sknet}.\par
\begin{figure}[t]
	\begin{center}
	    \includegraphics[width=0.75\textwidth]{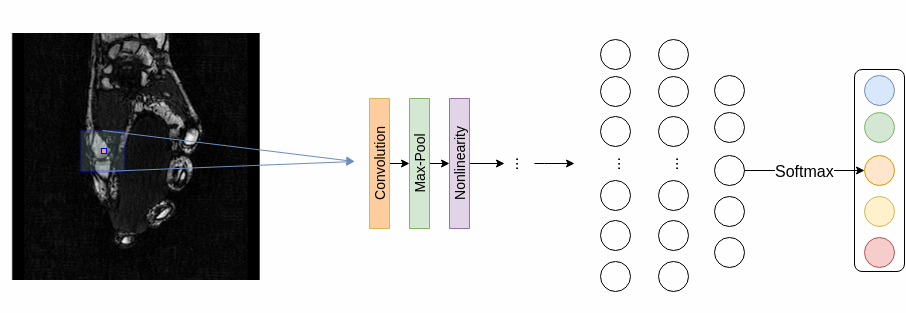}
	\end{center}
	\caption{A typical classifier network working on a 2D slice of a hand MRI. The patch marked in blue is input to the network to classify its central pixel. Notice how in order to classify nearby pixels the network will have to process near identical patches. The hidden layers can be substituted by convolutions to obtain an FCN.}
\end{figure}
One feature of classifier deep neural networks that is limiting in a segmentation setting is the usage of fully connected layers. Figure 1.1 depicts a typical CNN consisting of multiple convolution and pool layers followed by a series of fully connected layers that produce the final output for a 5-way classification problem. Using fully connected layers to produce multi-pixel outputs is only straightforward in a binary segmentation setting, where the class of a pixel can be determined by applying a threshold to the output neuron representing it. An approach first proposed by Lin et al.\ \cite{nin} is to remove fully connected layers from CNNs. Long et al.\ \cite{fcn} later proposed their replacement with convolutions using $1\times1$-sized filters. A network modified in this way can undertake multi-class prediction for multiple pixels in a single forward pass if the size of the input is larger than the total size reduction applied by the convolution and pool layers in the network \cite{fcn}.\par
While replacing fully connected layers with convolutions allows for multiple pixels to be predicted efficiently and simultaneously, the resulting outputs typically have a very low resolution. This is caused by the fact that CNNs reduce the size of their input in multiple ways. Convolution layers with filters larger than $1\times1$ reduce the size of their input. This loss of pixels can be prevented by extending the input with a padding of zero-valued pixels. Max-pooling layers or strided convolutions are also often-used building blocks of CNNs that cause a high size reduction. Both of these layers involve applying a filter to small windows of an input feature map separated by a number of pixels $k > 1$. In the case of max-pooling, this filter only retains the maximum of each window. The main motivation of using such operations in a network tasked with segmentation is to increase the network receptive field, allowing each output neuron to see a larger portion of the input. In addition, downsampling allows for a higher number of feature maps to be used without overwhelming system memory, which is especially of concern in a 3D image data task. \par
Different methods have been proposed to tackle the problem of creating a CNN that can generate a segmentation map for an entire input image in a single forward pass. Long et al.\ \cite{fcn} restore the downsampled feature maps to the original size of the input using deconvolution operations. Deconvolution is proposed by Zeiler and Fergus \cite{zeiler} as a means of visualizing CNNs. The authors of \cite{fcn} refer to it as backwards-strided convolution, using learnable kernels that are initialized to compute bilinear interpolation. The coarse segmentation map acquired through deconvolution is made finer by combining it with segmentation maps computed from earlier stages in the network, where the input has been downsampled fewer times. The authors refer to their network design as a ``fully convolutional network'' (FCN). \par
An alternative way of upsampling feauture maps is used by Badrinarayanan et al.\ \cite{segnet} and Noh et al.\ \cite{deconvnet}. These authors store the indices of activations that were preserved by max-pooling layers. The indices are later used for upsampling. Unpooling feature maps in this fashion creates sparse feaure maps, in which all locations but the ones previously selected by max-pooling contain zero-values. Further convolutions can turn these into dense feature maps. This method of unpooling was first introduced by Zeiler et al.\ \cite{zeiler2}.\par
Some authors have tried coupling a CNN with a ``Conditional Random Field'' (CRF) to enhance segmentation results. The CRF is used to model the relationship between pixels or regions in an image and is often used as a stand-alone segmentation method \cite{fccrf,lscrf,verbeek}, as well as in conjunction with a CNN \cite{cnncrf,deconvnet,spconv}. In the domain of medical image segmentation, Kamnitsas et al.\ \cite{kamnitsas} let the output of a CNN be postprocessed by a CRF. In \cite{braint}, it is proposed to train a second CNN to fulfill the functionality of a CRF. The authors point out the efficiency gain of substituting the computationally expensive CRF with a CNN.\par
Successful CNN-based medical image segmentation methods often draw on these recent findings in semantic segmentation. The two most common approaches are training a CNN on patches extracted from images and doing inference by sliding the CNN across all pixels of the network, predicting one pixel in each forward pass \cite{ciresan,pereira} and training an FCN on full images or large sections of images \cite{unet,3dunet,vnet,voxresnet,liver,kamnitsas,braint}. Certain aspects of medical images call for modifications on designs that are created for semantic segmentation, most notably the increased memory demands of 3D images and the imbalance between labels found in 3D ground truth data. These will be further discussed in the next section.

\subsubsection{Applications of CNNs to Medical Image Segmentation}
\label{subsection:MedCnns}
An early application of Deep Neural Networks to medical image segmentation is reported by Cire\c{s}an et al.\ \cite{ciresan}. Here, the authors take on the task of segmenting stacks of electron microscopy (EM) images, using a 2D convolutional neural network. To segment an entire stack, the classifier is applied to each pixel of every slice in a sliding window manner by extracting a patch around the pixel. This introduces the previously discussed two drawbacks: time-inefficiency caused by redundant computation and the inability of the network to learn global features. It is also noted in \cite{ciresan} that segmenting an entire stack with their approach takes ``10 to 30 minutes on four GPUs'' and that training the network on larger patches yields better results, supporting the above two statements. It is also important to note that these images are relatively small in the third dimension, with a size of $512\times512\times30$. This lack of depth allows them to be processed slice-by-slice as 2D images.\par
\begin{figure}[t]
	\begin{center}
	    \includegraphics[width=0.75\textwidth]{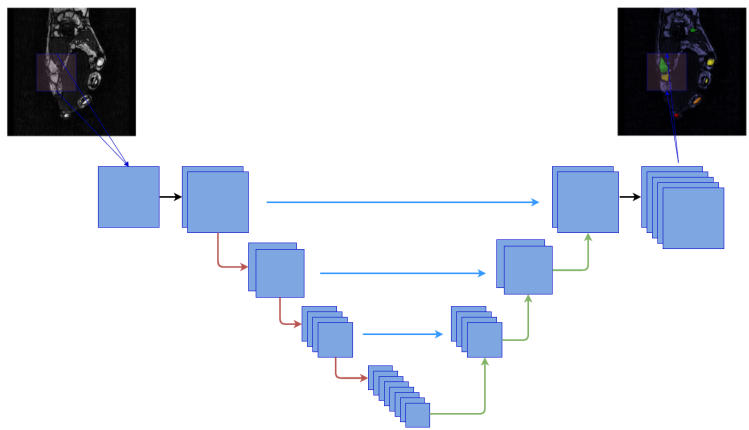}
	\end{center}
	\caption{A ``U-Net''-like architecture in abstract form. The number of feature maps depicted is symbolic. Arrows marked red are blocks of layers that downsample their input once, while green arrows do upsampling and black ones keep the original size of their input. The cyan arrows are long skip connections concatenating feature maps from the first stage to feature maps in the second stage, which means the expanding path has access to twice as many features as the contracting path.}
\end{figure}
Ronneberger et al.\ \cite{unet} attempt the same task with better results, using the fully convolutional approach of \cite{fcn} discussed in the previous section. Their design is similar to that of a convolutional autoencoder: the network consists of a ``contracting'' and an ``expanding'' stage. In the former, the size of the input is reduced while the number of feature maps are increased; in the latter the reverse is true. The abstract form of their architecture is depicted in Fig. 1.2. One important building block are skip connections which forward feature maps computed in the contracting stage to the expanding stage. These represent a direct flow of feature maps from an early layer in the network to a later one and recur in many works focusing on segmentation tasks. In the case of \cite{unet}, they are realized by concatenating the feature maps of the early layer to those of the later one, though a more common approach is to use element-wise summation \cite{resid, fcn}. A convolution with filters of size $1\times1$ following the last layer of the expanding stage produces the final segmentation. For segmentation, an image is separated into regions that are classified by the network sequentially and their respective outputs are used to tile the final segmentation. This decision has the advantage that the network can segment arbitrarily large images by dividing them into regions, while a network attempting to segment entire images would not be able to process images exceeding a certain resolution due to memory constraints. The authors have called their 2D convolutional neural network ``U-Net''.\par
The U-shaped architecture of Ronneberger  \cite{unet} was extended with 3D convolutional kernels by \c{C}i\c{c}ek  \cite{3dunet} and Milletari  \cite{vnet}. This extension is necessary, as the original 2D architecture is only suitable for stacks of 2D images or 3D images with low depth. A large portion of medical images, such as scans of the brain and other organs, have comparable sizes in each dimension, making slice-by-slice application of a 2D network inefficient. Milletari et al.\ \cite{vnet} have further modified the original U-Net design by the usage of residual blocks. First introduced in \cite{resid}, residual blocks make use of special additive skip connections to combat vanishing gradients in deep neural networks. At the beginning of a residual block, the data flow is separated into two streams: the first carries the unchanged input of the block, while the second applies weights and non-linearities. At the end of the block, the two streams are merged using an element-wise sum. The main advantage of such constructs is to allow the gradient to flow through the network more easily. Figure 1.3 depicts the original residual block design \cite{resid} along with a more recent proposal \cite{resid2}.\par
\begin{figure}[t]
	\begin{center}
	    \includegraphics[width=0.75\textwidth]{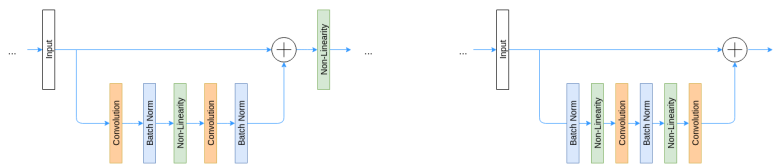}
	\end{center}
	\caption{Original (left) and more recent (right) residual block designs. Note that on the right, the gradient can flow uninterruptedly since no non-linearity is applied.}
\end{figure}
The previously discussed autoencoder-like architectures U-Net \cite{unet}, 3D U-Net \cite{3dunet} and V-Net \cite{vnet} have several important similarities. First, all three networks are trained on entire images or larger sections of images as opposed to small patches. As a consequence, they have to deal with two limitations posed by the task of medical image segmentation: data scarcity and class imbalance. Creating 3D segmentation ground truths for medical images requires time and training. As a result, most medical image data sets for segmentation tasks contain very few samples. While previous methods could rely on sampling patches to create large data sets, these new networks need other ways of increasing the effective number of samples they can train on. In all of the three cited papers, the authors report using extensive image transformations for training, which they claim to be vital for successful generalization. The transformations used by \cite{unet,3dunet,vnet} include shifting, rotating and scaling images, as well as augmenting grey values. One transformation cited in each of the three papers is the application of a random deformation field. This random deformation field is applied to the image at the beginning of each training iteration. Hence, in every epoch, the network is training on a different version of the original data set.\par
The switch from patch-based training to pixel-wise training on full images requires some special care for class imbalance. While methods relying on patch based training can keep a balanced data set by sampling an equal number of patches belonging to each class, data sets of entire images are bound to the original distribution of classes. In the case of medical images, this distribution is dominated by negative (``background'') voxels. Especially in a setting with one background and several foreground classes (such as different types of bones or regions of tumorous tissue), class imbalance becomes overwhelming. The solution used by Ronneberger et al.\ \cite{unet} and \c{C}i\c{c}ek et al.\ \cite{3dunet} is applying a weight map to the categorical cross-entropy loss function, while Milletari et al.\ \cite{vnet} seek to maximize the dice similarity coefficient instead. An obvious advantage of the latter is that it does not rely on any hyperparameters.\par
Skip connections prove themselves useful in U-Net, 3D U-Net and V-Net and they are also found in residual networks, though here they are used in summation and not concatenation. The benefits of skip connections are investigated by Drozdzal et al.\ \cite{skip} in a medical image setting. They make a distinction between long skip connections arching over a large number of layers (such as the feature forwarding connections between the contracting and expanding stages of U-Net) and short skip connections spanning over a single residual block (which usually contains no more than two weight layers). Both types of skip connections are found to be beneficial in creating deep architectures for medical image segmentation.\par
Other CNN-based medical image segmentation methods following the FCN approach are reported by Chen et al.\ \cite{gland,voxresnet} and Dou et al.\ \cite{liver}. The prevalent approach in these works is to omit the expanding path found in U-Net-like architectures and instead create low resolution segmentation maps that are then deconvolved to the original resolution and combined to produce the final segmentation. Another method seen in \cite{liver} and \cite{voxresnet} is ``deep supervision''. Here, feature maps from earlier points in the network are used to create secondary segmentation maps. These are used in two ways: They are combined with the final segmentation, and the losses associated with these segmentation maps are weighted and added to the final loss function.\par
Kamnitsas et al.\ \cite{kamnitsas} introduce a 3D CNN architecture designed for various segmentation tasks involving MR images of brains. The authors benchmark their approach on the BRATS \cite{brats} and ISLES \cite{isles} challenges\footnote{Refer to http://www.braintumorsegmentation.org/ and http://www.isles-challenge.org/, respectively.}. Their approach comprises a CNN with 3D filters and a conditional random field smoothing the output of the CNN. Similar to \cite{unet}, the authors propose dividing the input images into regions in order to address the high memory demand of 3D CNNs. Notable in \cite{kamnitsas} is the usage of an architecture consisting of two pathways. The first receives the subregion of the original image that is to be segmented, while the second receives a larger region that is downsampled to a lower resolution before being fed to the network. This enables the network to still be able to learn global features of the images.\par
A similar approach is followed by Havaei et al.\ in \cite{braint}, also training and testing MR images of brains from the BRATS and ISLES data sets. The authors process the images slice-by-slice using 2D convolutions. Also, Havaei et al.\ use the second part of their two-path architecture to fulfil the functionality of a CRF by feeding the segmentation output of the first path to the second one. Like in \cite{kamnitsas}, the first CNN receives a larger portion of the original image than the second, with the purpose of learning both global context and local detail.\par
To conclude, the variety of CNN-based medical image segmentation methods is largely due to different attempts at addressing difficulties specific to medical images. These are chiefly the memory demands of storing a high number of 3D feature maps, the scarcity of available data and the high imbalance of classes. In dealing with the first issue, most researchers have turned to dividing images into a small number of regions and stitching together the outputs of different regions \cite{unet,kamnitsas,voxresnet} and/or using downscaled images \cite{3dunet}. Data augmentation is often used to address the scarcity of data \cite{vnet,3dunet,unet,ciresan,kamnitsas,skip}. As for class imbalance, reported methods include using weighted loss functions \cite{unet,3dunet}, overlap metrics such as the dice similarity \cite{vnet,skip} or deep supervision \cite{voxresnet,liver}.

\section{Method}
\label{chapter:Method}

\subsection{Network Architecture}
Following the example of other successful architectures discussed in the previous section, the ``fully convolutional'' approach is used in this work. Figure 2.1 depicts the baseline for the models tested. The architecture consists of the contracting and expanding stages found in other works using similar networks \cite{unet,3dunet,vnet}, where feature maps from the first stage are combined with feature maps from the second stage via long skip connections. The prevalent approach in combining feature maps in other works has been concatenation. In this work, concatenation is compared with element-wise summation\footnote{The CUMED-team have reported using this method in their winning submission to the Promise 12 challenge, a manuscript of which can be viewed under \url{https://grand-challenge.org/site/promise12/Results/}}. \par 
Element-wise summation directly inserts local details found in the feature maps of the contracting stage to the feature maps of the expanding stage. In addition to this, all of the layers between the source and the destination of the skip connection, together with the skip connection itself, can be viewed as one large residual block. Existing literature on residual networks has theorized that residual blocks are most useful when the optimal function approximated by a stack of layers in a deep neural network is close to the identity function \cite{resid}. Hence, feature maps created by the expanding stage act as refinements of feature maps created in the contracting stage. Finally, the network is also tested without any usage of the long skip connections in order to judge their usefulness.\par
\begin{figure}[t]
	\begin{center}
	    \includegraphics[width=1.\textwidth]{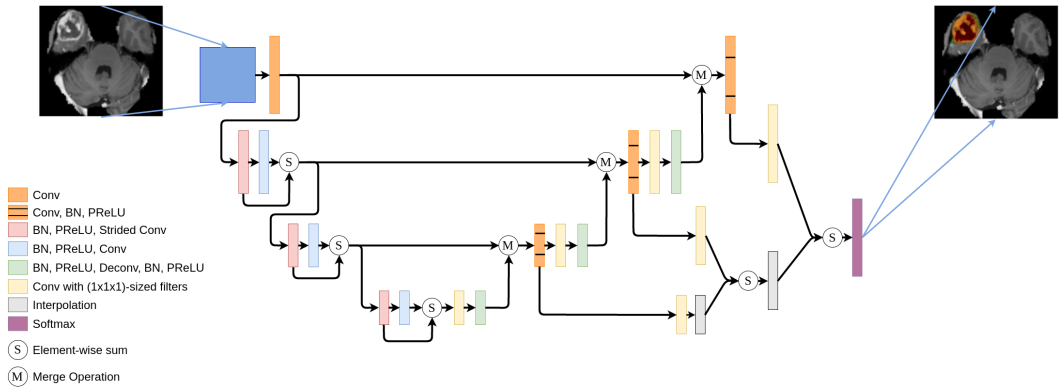}
	\end{center}
	\caption{The basic architecture used throughout the experiments. The actual architecture is 3D. 2D is used here for the purpose of simplicity. Two different merge operations are tested for the long skip connections: element-wise addition and feature concatenation. Best viewed in electronic format. \label{fig:network}}
\end{figure}
Furthermore, this work looked at combining segmentation maps created at different points in the network. The idea was used in the original FCN-design by Long et al.\ \cite{fcn} as a means to reduce the coarseness of the final segmentation. Since then, other works on 3D medical image segmentation have reported creating multiple segmentation maps at different resolutions, which are then brought to a matching resolution by deconvolution and combined via element-wise summation \cite{voxresnet,liver}. As depicted in Fig.~2.1, three segmentation maps are created by the network used in this work: one having the same size as the input, one having one half the size in each dimension and one having one fourth the size in each dimension. These are then combined in the following way: First the segmentation map with the lowest resolution is upsampled with bilinear interpolation to have the same size as the second-lowest resolution segmentation map. The sum of the two is then upsampled and added to the highest-resolution segmentation map in the same way. \par 
Since the expanding path already receives local details from multiple long skip connections, combining the final segmentation produced at the last layer of the network with additional segmentation maps does not have the purpose of refinement. Rather, it is used as a means to speed up convergence by ``encouraging'' earlier layers of the network to produce good segmentation results. Similar techniques are used in \cite{inception,dsn,voxresnet}. Results obtained with and without the usage of this method are provided in Section~\ref{chapter:Results}. \par
As discussed in the previous section, the high memory demand of 3D medical images poses a difficulty. A common solution used in a number of publications is to divide the images into a small number of sections during inference, which are then segmented sequentially and stitched together \cite{unet,voxresnet,3dunet}. When this method is chosen, training is also done on large sections sampled from the images \cite{unet,voxresnet,3dunet,kamnitsas}. An alternative is to use downsampled images whenever the memory demand becomes overwhelming. With this approach, the output of the network is a low-resolution segmentation, which can be upsampled using interpolation or superresolution. This is the training and inference scheme used in this work for images that are too large to be processed by the network. Section~\ref{chapter:Experiments} goes into further detail on this point. \par
Throughout the network, zero-padding is used so that the size of feature maps is only changed by strided convolutions or deconvolutional operations. The network uses PReLU activations \cite{prelu}, which are defined as: $f(x)=\max\{0,x\} + a\min\{0,x\}$; $a$, in this case, is a learned parameter. Each activation layer is preceded by a layer of batch normalization \cite{bn}. Strided convolution is used in place of max-pooling as it was found to yield slightly better results in the preliminary experiments. The suggestion to replace max-pool layers with convolutions is also made by Springenberg et al.\ \cite{nopool} and found to be useful in a medical image segmentation setting by Milletari et al.\ \cite{vnet}. All convolutions used in the network have filters of size $3\times3\times3$, with the exception of the final convolutions used to produce the segmentation maps and the ones involved in reducing the number of feature maps prior to a deconvolutional layer. Both of these use $1\times1\times1$-sized filters. Further specification of the network can be found in Appendix~\ref{app:A}. Details about the receptive field of the network are included in Appendix~\ref{app:B}.
\subsection{Loss Metric}
As stated in Section~\ref{chapter:Intro}, class imbalance is a serious issue in medical image segmentation and one that requires special care with regard to the loss function used. Inspired by \cite{vnet}, a loss function close to the dice similarity coefficient is used in this work for training. The dice coefficient is an overlap metric frequently used for assessing the quality of segmentation maps. It is defined as:
\begin{equation}
    \mathrm{DSC}=\frac{2|P \land T|}{|P| + |T|} 
\end{equation}
Where P and T are two sets. The formulation for bit vectors would be:
\begin{equation}
    \mathrm{DSC}=\frac{2 \lVert PT \rVert _{2}}{ \lVert P \rVert _{2} ^{2} + \lVert T \rVert _{2} ^{2}} 
\end{equation}
Where $PT$ is the element-wise product of $P$ and $T$ and $\lVert X \rVert _{2}$ is the L2-Norm of $X$. The loss function used throughout the experiments is based on a similarity metric that is formulated very closely:
\begin{equation}
\label{eq:jacc}
    \mathrm{Jacc}=\frac{\lVert PT \rVert _{2}}{\lVert P \rVert _{2} ^{2} + \lVert T \rVert _{2} ^{2} - \lVert PT \rVert _{2}} 
\end{equation}
This metric is widely known as the Jaccard Index. It can be shown that: $\mathrm{DSC}=2\mathrm{Jacc}/(1+\mathrm{Jacc})$. Instead of maximizing the Jaccard Index, the network is trained to minimize the Jaccard Distance defined as: $1-\mathrm{Jacc}$. It has the following intuitive formulation:
\begin{equation}
    \mathrm{Loss}=\frac{\text{False Positives}+\text{False Negatives}}{\mathrm{Union}}
\end{equation}
Where ``Union'' refers to the set of voxels labelled as positive in either the prediction or the target segmentation map or both. The final loss is accumulated by taking the sum of $1-\mathrm{Jacc}$ for each class. \par
It should be noted that this choice of loss function has some limitations. In the case of samples for which the ground truth contains no foreground labels (e.g. a brain MRI scan containing no tumorous regions), the loss will be maximized, even if the output segmentation map contains very few false positives. One consolation is that the loss is bounded between 0 and 1, meaning the loss cannot blow up. Still, it must be taken into account that the loss function is poorly defined for samples containing very few or no instances of foreground classes. Another limitation stems from the fact that both the dice coefficient and the Jaccard index are only defined for binary maps. One extension of the Jaccard index to multi-class segmentation maps is defined as:
\begin{equation}
    \mathrm{Jacc}=\frac{\sum_i \min\{P_i, T_i\}}{\sum_i \max\{P_i, T_i\}}
\end{equation}
Where $X_i \in \{c_0,\dots,c_n\}$ for discrete labels $c_j$. However, one problem of this definition is that some forms of misclassification will be penalized more severely than others. This is true for three labels $c_0$, $c_1$ and $c_2$ with $c_2 > c_1 > c_0$. In this case, misclassifying a voxel for which the true label is $c_2$ with the label $c_0$ will yield a higher loss than misclassifying it with the label $c_1$. In order to avoid this, the definition in Eq.~(\ref{eq:jacc}) is used instead, applied separately for each class. This decision means the network is trained on multiple, one-vs-all binary classification tasks simultaneously, as opposed to one multi-class classification problem. It is noted that such training schemes may produce ambiguously classified regions in input space \cite{bishop}.
\section{Experiments}
\label{chapter:Experiments}

\subsection{Data}
\label{section:Data}
\begin{figure}
	\begin{center}
	    \includegraphics[width=0.5\textwidth]{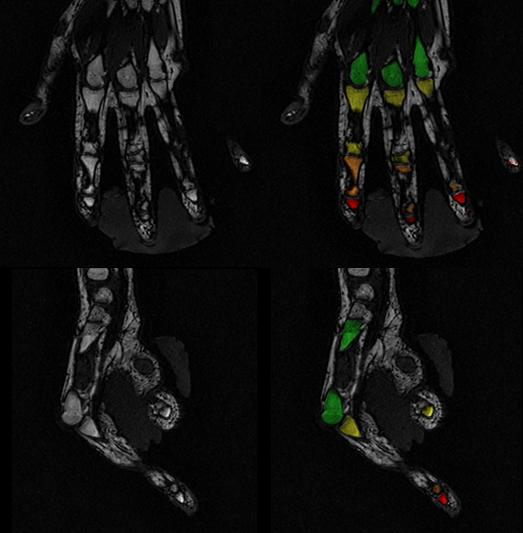}
	\end{center}
	\caption{2D slices of images belonging to the hand MRI data set. The ground truth comprises five classes: metacarpal phalanx (green), proximal phalanx (yellow), middle phalanx (orange) and distal phalanx (red). The visible gap in the proximal phalanx of the index finger is caused by the fact that the middle section of the bone is not visible in the original image. \label{fig:hand_mri}}
\end{figure}

The network introduced in the previous section was applied to two different tasks. The first concerns the segmentation of bones in hand MRI. The data set for this task comprises 30 MRI volumes of a hand in different postures \cite{Gustus}. Figure~\ref{fig:hand_mri} shows 2D slices from two images belonging to the data set, with and without ground truth labels. The original size of the images is $480\times480\times280$. The following labels are included in the ground truth: (1) metacarpal phalanx, (2) proximal phalanx, (3) middle phalanx and (4) distal phalanx. Since forward and backward pass of the images in their original size would require more memory than available, they are downsampled to one fourth the size in each dimension using bilinear interpolation (amounting to a volume reduction by factor 64). The downsampled images are then cropped along the z-Axis so as to discard regions depicting empty space above and below the hand, resulting in a size of $120\times120\times64$. \par
\begin{figure}
	\begin{center}
	    \includegraphics[width=0.5\textwidth]{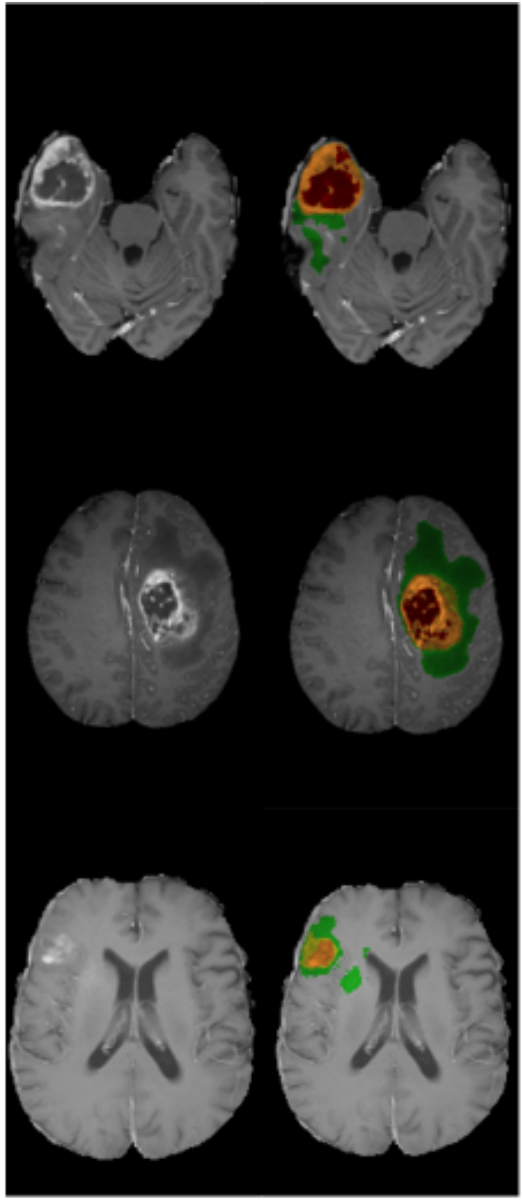}
	\end{center}
	\caption{2D slices of three images from the BRATS 2015 data set. Raw images (left) and overlayed ground truth (right). Colors correspond to: necrosis (red), enhancing tumor (orange), non-enhancing tumor (yellow), edema (green). The images are in the T1C modality. \label{fig:brain_mri}}
\end{figure}

The second task is the main objective of the annual BRATS challenge: segmenting tumor regions in brain MRI. More concretely, the network is trained using the BRATS 2015 training set, which contains a total of 274 images in four modalities: Flair, T1, T1C and T2. Of the 274 images, 220 were classified as high grade gliomas, while the remaining 54 were classified as low grade gliomas, with no inclusion of images depicting healthy brains\footnote{This is a limitation addressed in Section~\ref{chapter:Results}.}. The classes involved in the segmentation task are: (1) necrosis, (2) edema, (3) non-enhancing and (4) enhancing tumor. Several examples are depicted in Fig~\ref{fig:brain_mri}. All of the images have a size of $240\times240\times155$, which can be cropped to a region of the size $160\times144\times128$, while still containing the entire brain. For some of the experiments, these cropped images were further downsampled to a size of $80\times72\times64$ for training. For the final submission to the online evaluation tool of the BRATS Challenge, training and inference were done on the original $160\times144\times128$-sized cropped images. \par
In order to demonstrate the severity of class imbalance in these segmentation tasks, the average frequency of each class taken over the entire data set was calculated for both the hand and the brain MRI. Table~\ref{tab:mean_freqs} collects these metrics. In both data sets the background classes have a very high frequency ($\approx$ 96\% and $\approx$ 99\%). In the set of hand MRI, the distal and middle phalanxes are nearly one order of magnitude less frequent than the other foreground classes due to their smaller sizes. Class imbalance is less severe in the case of the brain MRI, since cropping has removed large regions of empty space. This was not possible for the hand MRI because the varying postures assumed by the hand do not adhere to a common and meaningful bounding box size.\par
\begin{figure}
	\begin{center}
	    \includegraphics[width=0.75\textwidth]{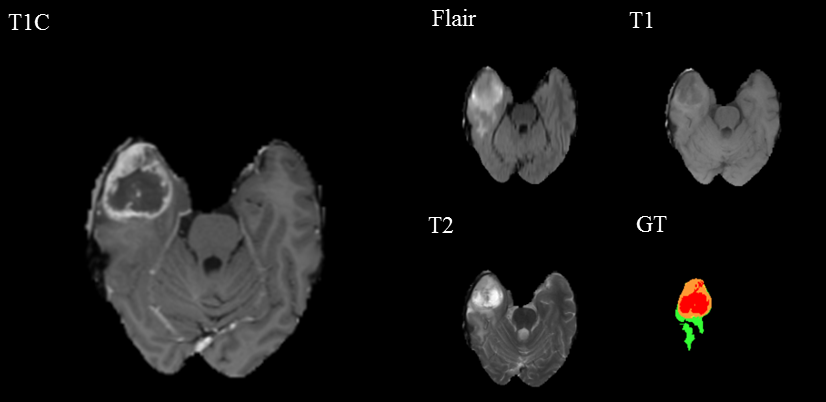}
	\end{center}
	\caption{The different modalities available for each image, together with ground truth labels. T1C is enlarged here for a better view of the details. Colors correspond to: necrosis (red), enhancing tumor (orange), non-enhancing tumor (yellow), edema (green). \label{fig:mods}}
\end{figure}

\begin{table}
\centering
\caption{Average class frequency taken over all images for each data set, approximate values. Labels are: background (0), metacarpal/necrosis (1), proximal/edema (2), middle/enhancing tumor (3) and distal/non-enhancing tumor (4). \label{tab:mean_freqs}}
\begin{tabular}{c c c c c c}
\multicolumn{1}{c}{ } & \multicolumn{5}{c}{Average class frequency} \\
\cline{2-6}
\multicolumn{1}{c}{ } & 0 & 1 & 2 & 3 & 4 \\
\hline
Brain & 0.96 & $0.002$ & $0.024$ & $0.005$ & $0.006$ \\
\hline
Hand & 0.99 & $5.13*10^{-3}$ & $2.29*10^{-3}$ & $6.78*10^{-4}$ & $4.32*10^{-4}$ \\
\end{tabular}
\end{table}
It is worth noting that these two tasks are significantly different in their difficulty. The anatomy of the hand poses limitations on the shape, size and location of the bones. Tumors, on the other hand, can manifest at vastly different locations, shapes and sizes \cite{dlbrtrends}. In order to capture different regions of interest, several different MRI modalities are used. These are collected in Fig.~\ref{fig:mods}. Images in the Flair and T2 modalities are useful for detecting edema, while T1C can be used to detect inner regions of the tumor (necrosis and enhancing core) and T1 can assist in the separation of the tumor core from edema, according to other reports \cite{brats,dlbrtrends}. \par
Menze et al. \cite{brats} have compared segmentation maps created by different human experts using the dice similarity coefficient in order to set a benchmark for automated methods. They compare results on three principal regions: ``whole'' (separation of healthy voxels from non-healthy ones), ``core'' (separation of the tumor core consisting of necrosis, enhanced and non-enhanced tumor regions from edema and healthy tissue) and ``active'' (separation of enhanced tumor regions from the rest of the classes). When two segmentation maps created for the same image by two different human experts are compared, the average agreement rates are 85\%, 75\% and 74\% for the ``whole'', ``core'' and ``active'' regions respectively, based on the dice similarity coefficient \cite{brats}. When, instead, a segmentation map created by one human expert is compared to the consensus segmentation of multiple experts, the average dice scores increase to 91\%, 86\% and 85\%. The consensus results are produced by taking multiple segmentation maps (each one created by a different human expert) and fusing them together using an algorithm \cite{brats}; thereby generating a more reliable segmentation. These consensus segmentations are also used as ground truth data for supervised methods such as the CNNs described in this work to train on. \par

\subsection{Learning and Evaluation}
\subsubsection{Training}
Subsection~\ref{subsec:exps} enumerates the various experiments conducted on each data set. The following describes the general training set-up used throughout the experiments: Of the 30 MRI volumes in the hand data set, 20 are used for training, while the remaining 10 are reserved for the validation and test sets, with 5 images each. In order to address the scarcity of the data available, random transformations are applied to the images. At each training iteration one of three transformations is selected to be used on the image that is to be classified by the network. These are the identity function (so as to ensure the network can see the original images as well), flipping along a random axis, and rotation by a random angle along two randomly selected indices. The network is trained using the Adam optimization algorithm \cite{adam} on an NVIDIA Tesla k40c GPU for 600 epochs (approximately 24 hours) or until validation loss is unchanged for 100 epochs. The learning rate is set to $5*10^{-5}$, while $\beta_1$ and $\beta_2$ are set to 0.1 and 0.001 respectively. The trained network takes roughly 0.9 seconds to segment one $120\times120\times64$-sized image. \par
Training on the BRATS data set is set up similarly. The 274 images in the data set are partitioned into the training, validation and test sets with 220, 27 and 27 images respectively. Random transformations are used in the same way as on the hand images. The Adam optimizer is used with the same settings as for the hand data set. Training takes place on an NVIDIA Tesla k40c for 50 epochs (around 62 hours).  The trained network takes roughly 3 seconds to segment one $160\times144\times128$-sized image. \par
In all of the reported results, network weights were initialized by sampling from a Gaussian distribution with zero mean and a standard deviation of 0.01. The initialization method commonly known as ``Xavier initialization'' \cite{xavier}, where the parameters of each layer are sampled from a Gaussian with zero mean and unit variance, and scaled by a factor of $\sqrt{2/(n_\mathrm{in} + n_\mathrm{out})}$ \footnote{$n_\mathrm{in}$ and $n_\mathrm{out}$ refer to the number of inputs and outputs of the layer.} was tested on the hand data set but found to produce worse results.
\subsubsection{Cross-Validation}
As stated before in Section~\ref{chapter:Method}, this work compares the usage of element-wise summation with the usage of concatenation when feature maps from one stage of the network are to be combined with feature maps from the other. In the case of the hand data set, the dice score achieved for each class averaged over 6 cross-validation folds is reported for both the summation and the concatenation network. For the BRATS data set, two different sets of evaluation results are reported: first the average dice score for each region taken over 5 cross-validation folds, and second, an official evaluation using 110 additional, unlabelled images. The latter is provided by the BRATS organizers along with the 274 labelled training images. Evaluation is done by uploading the segmentation maps created by the network to an online portal. 
\subsubsection{List of Experiments}
\label{subsec:exps}
Experiments on the hand data make comparisons between:
\begin{enumerate}
\item The Jaccard loss and categorical cross-entropy\footnote{For a definition, see Section~\ref{chapter:Results}.}
\item A network with long skip connections and a network without
\item A network combining multiple segmentation maps and a network creating a single segmentation
\item Element-wise summation and concatenation
\end{enumerate}
Where numbers 3 and 4 are conducted on 6 cross-validation folds. The next section goes into more detail on all of these points. Experiments involving the BRATS data set look at the following:
\begin{enumerate}
\item Performance achieved using each modality on its own
\item Performance achieved using different combinations of the modalities
\item Final performance achieved by the best-performing network
\end{enumerate}
Of these, point 3 takes place over 5 cross-validation folds using the original non-downsampled images, while the points 1 and 2 are done using downsampled images without cross-validation in order to speed up the process. The network architecture used in point 3 was selected according to the experiments conducted on the hand data set.
\section{Results}
\label{chapter:Results}
\subsection{Results Obtained on Hand MRI}
\label{section:HandRes}
\begin{figure}
	\begin{center}
	    \includegraphics[width=0.75\textwidth]{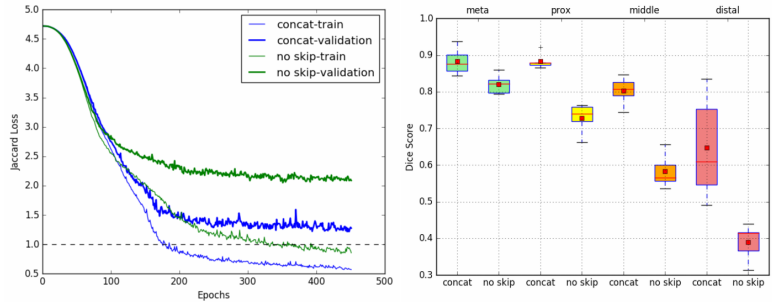}
	\end{center}
	\caption{Comparison of a network with feature forwarding long skip connections and a network without. The plot on the left compares the learning curves of the two models, while the box plot on the right details the test set dice scores reached on the four bone classes: metacarpal (green), proximal (yellow), middle (orange), distal (red). \label{fig:conc_vs_none}}
\end{figure}

An important part of the network described in Section~\ref{chapter:Method} are long skip connections forwarding feature maps from the contracting stage to the expanding stage. These allow the network to recover local details which are lost due to the usage of downsampling operations. In order to justify the inclusion of long skip connections, the network was trained and tested on the hand MRI data set described in Section~\ref{chapter:Experiments} with and without long skip connections. The results of this experiment are summarized in Fig. \ref{fig:conc_vs_none}. It was revealed that removing long skip connections unambiguously worsens the performance of the network.\par
\begin{figure}
	\begin{center}
	    \includegraphics[width=0.45\textwidth]{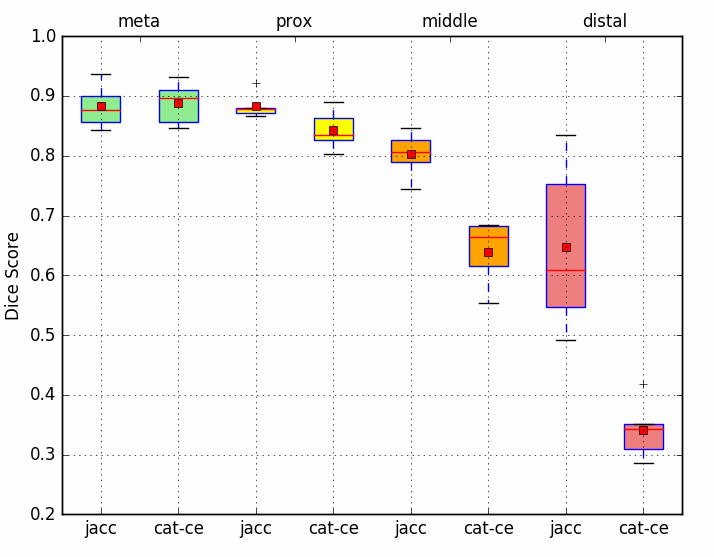}
	\end{center}
	\caption{Comparison of a network trained with the Jaccard distance with a network trained with categorical cross-entropy, based on test set dice scores. The network trained with categorical cross-entropy proves far less capable of detecting infrequent classes. \label{fig:jacc_vs_cat_ce}}
\end{figure}

As stated before in Section~\ref{chapter:Method}, the Jaccard distance was used as a loss function in this work. A similar approach to address class imbalance was taken in \cite{vnet}. In Fig.~\ref{fig:jacc_vs_cat_ce}, a network trained with the Jaccard distance is compared to a network trained with simple categorical cross-entropy, which is a loss function that corresponds to the maximum likelihood solution of a multiclass classification problem. It is defined as: $H(t,p)=-\sum t(x)\log (p(x))$, $p$ and $t$ corresponding to ``prediction'' and ``target''. As the dice scores reveal, both networks perform well on the first two classes, which are relatively common labels in the ground truth data. Turning to the least frequent two classes ``middle'' and ``distal'', it becomes clear that the network trained on categorical cross-entropy is far less capable of learning to detect highly infrequent classes, which can be explained by the fact that categorical cross-entropy corresponds to a maximum likelihood solution: the network gets biased towards more frequent classes, as this increases the likelihood of the training data. It was shown before in Table~\ref{tab:mean_freqs} that the classes ``middle'' and ``distal'' are the least common of all four foreground labels due to their smaller size. While the class imbalance could be countered by using weight maps wherein less frequent classes are associated with higher weights, this would introduce an additional hyper-parameter optimization problem. In order to avoid this, the Jaccard distance was used in all of the ensuing experiments.\par
\begin{figure}
	\begin{center}
	    \includegraphics[width=0.75\textwidth]{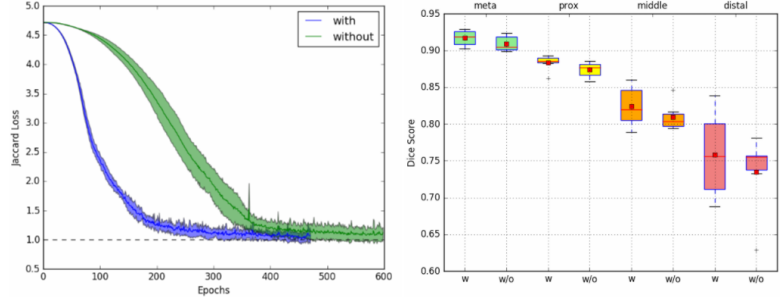}
	\end{center}
	\caption{Comparison of a network combining multiple segmentation maps (denoted as ``with'') and a network creating a single segmentation map (denoted as ``without''), based on the mean validation loss (left) and the mean test set dice scores (right) taken over six cross-validation folds. In the plot to the left, the area comprising one standard deviation is highlighted.\label{fig:with_vs_without}}
\end{figure}

Since the network used in this work differs from other, similar ``U-Net''-like architectures \cite{unet,3dunet,vnet} by the usage of multiple segmentation maps created at different sizes and combined via interpolation and element-wise summation (see Fig.~\ref{fig:network}), network performance without this addition was also tested in order to find out whether any improvement was achieved. Figure~\ref{fig:with_vs_without} draws a comparison between a network combining three segmentation maps, and one producing a single segmentation map at its very last layer. Both networks converge to similar values of the validation set loss, while the test set dice scores reveal the modified network to perform better. One faculty of the modified network that is clear to see is that it converges much faster, which is a desirable property when training time is the bottleneck. In light of these findings, the modification is kept throughout the experiments reported in this work. \par
\begin{figure}
	\begin{center}
	    \includegraphics[width=0.75\textwidth]{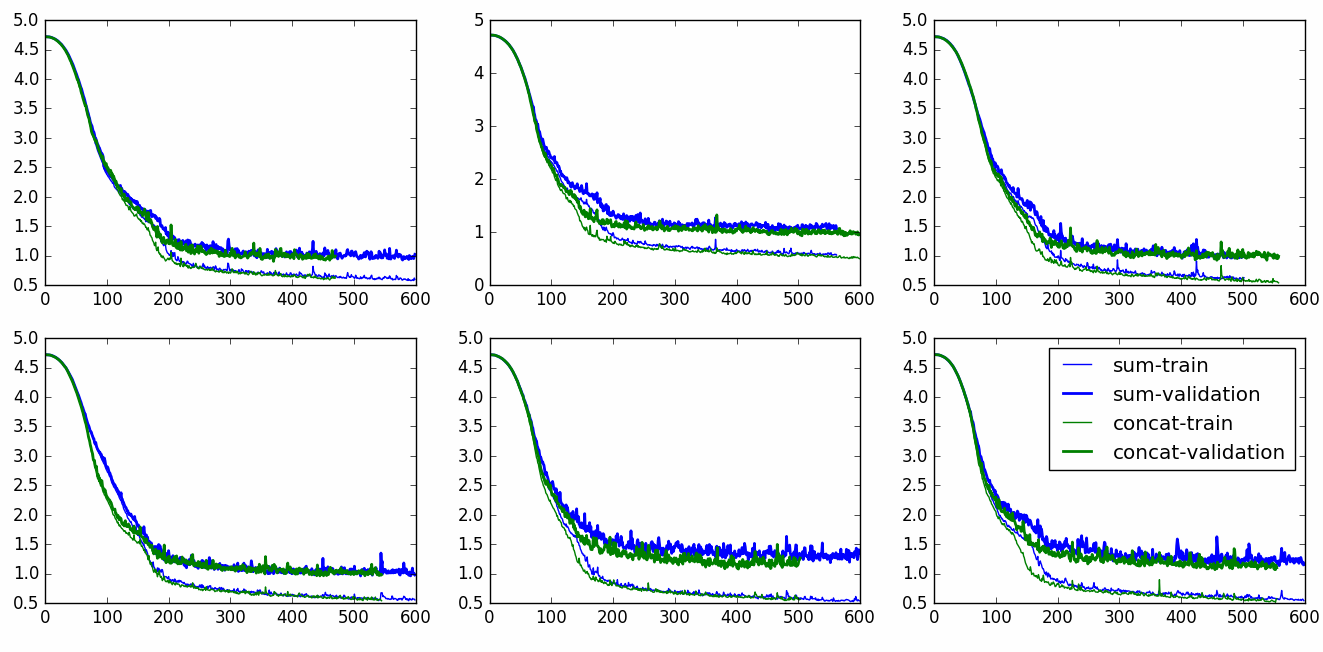}
	\end{center}
	\caption{Training and validation curves of the summation and concatenation networks for each cross-validation fold. Thick curves correspond to the validation loss. Because training was done until convergence, the curves vary in length. \label{fig:all_folds}}
\end{figure}

\begin{figure}
	\begin{center}
	    \includegraphics[width=0.75\textwidth]{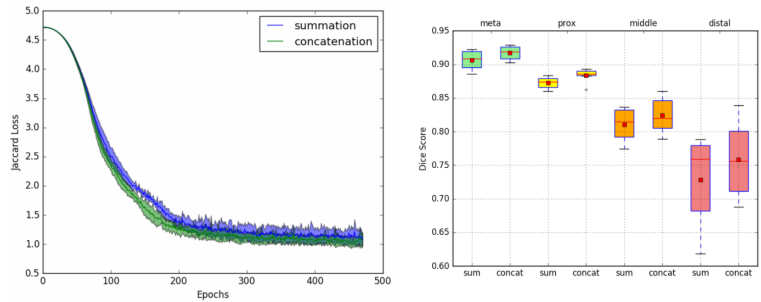}
	\end{center}
	\caption{The curves on the left belong to the validation loss achieved by each network throughout training and are averaged over six cross-validation folds. For each curve, the area comprising one standard deviation is highlighted. The box plot on the right details the dice scores obtained on each class over all cross-validation folds. \label{fig:val_test_comp}}
\end{figure}

Next, the network was trained and tested using two different types of long skip connections: concatenation and element-wise summation. Six cross-validation folds were used in order to get a sound comparison. The training and validation curves of all of the folds are collected in Fig.~\ref{fig:all_folds}.
The results suggest that the summation network is outperformed in all cross-validation folds. In order to confirm this, the test set dice scores obtained by the networks were averaged over all six cross-validation folds. These values, along with the curve of the validation loss achieved by each network averaged over all folds are visualized in Fig.~\ref{fig:val_test_comp}.
Like the single-fold learning curves, these plots suggest that concatenation works better than element-wise summation when used in long skip connections. \par
\begin{figure}
	\begin{center}
	    \includegraphics[width=0.75\textwidth]{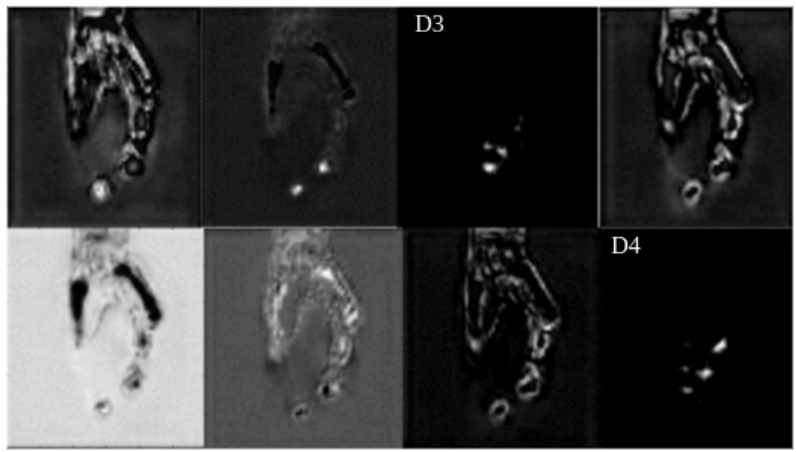}
	\end{center}
	\caption{Feature maps computed in the expanding stages of the concatenation and summation networks. The concatenation network produces early versions of clear segmentation maps (M1, M2, P, D1, D2), while the features of the summation network appear to include many irrelevant details (with the exception of D3 and D4). Please note that the feature maps are not shown in any specific order (i.e. the two sets of images are not meant to line up). Slices are extracted from differing locations in order to give a view of the distal phalanx. \label{fig:feats_sc}}
\end{figure}

In order to explain the superior performance of the concatenation network, feature maps from the expanding stages of both networks were visualized. These can be found in Fig.~\ref{fig:feats_sc}.
The feature maps visualized therein belong to the final portion of the expanding stage, where the features have been restored to the original size of the input images. By inspecting the images, it can be seen that the concatenation network produces features depicting global information (such as marking out the metacarpal and proximal phalanxes), alongside features depicting local details. Five of the eight features resemble final segmentation maps for different bone types: two seem to be marking the metacarpal phalanx (denoted as M1 and M2 in Fig.~\ref{fig:feats_sc}), another two the distal (D1 and D2) and a final one the proximal (P). The summation network, by comparison, has only produced two ``global-only'' feature maps (the two features marking the distal phalanx, denoted as D3 and D4). The rest of the feature maps all depict global information within a high amount of local detail. These are, for instance, the feature maps where areas corresponding to bones have been marked with black, while non-bone regions of the hand include many local details that are not relevant for the final segmentation. \par
The visualized features suggest that, while summing features from the expanding and contracting stages directly inserts local information into the feature maps, it also causes a high amount of irrelevant detail to be re-introduced into the information stream of the network, making it difficult for the network to selectively combine both types of information. By comparison, the concatenation network seems to benefit from keeping separate streams of information: one set of feature maps containing segmentation maps for the individual classes and another set containing local details.

\subsection{Results Obtained on the BRATS Data Set}
\label{section:BrainRes}
It was stated before in Section~\ref{chapter:Experiments} that the difficulties involved in segmenting gliomas motivate the usage of different MRI modalities. The images in the BRATS data set come in the four modalities Flair, T1, T1C and T2. In an attempt to assess the influence of each modality on the predictive performance of the network, the network was trained and tested on each one of the individual modalities separately. Based on the dice scores reached in these experiments, the network was then trained on a combination of the best performing modalities, where each modality represented one input channel. Table \ref{tab:mods} details the results obtained in this process. It was found that, when each modality is used on its own, T1C produces the best results on every class but edema, for which Flair, followed by T2, produces better predictions. Using this as a starting point, the network was then trained on Flair and T1C images together, which lead to dice scores very close to the ones achieved by a network that had access to all four modalities as input channels. Since T2 is the next best-performing modality, the network was then trained on images in the Flair, T1C and T2 modalities. Comparing the results of this run with the performance of the network when all modalities are available shows that, interestingly, the network achieves a much higher dice score on necrotic regions when the T1 modality is discarded. This suggests that some benefit could be achieved by training different networks on different combinations of MRI modalities. \par
\begin{table}
\centering
\caption{Dice similarity scores obtained by training on different MRI modalities. Note that the row ``all'' does not contain the best result obtained in this work. \label{tab:mods}}
\begin{tabular}{|r|c|c|c|c|}
\cline{2-5}
\multicolumn{1}{c|}{ } & necrosis & edema & enhancing & non-enhancing \\
\hline
 Flair & 0.19 & \textbf{0.74} & 0.28 & 0.36 \\
\hline
 T1 & 0.27 & 0.56 & 0.26 & 0.4 \\
\hline
 T1C & \textbf{0.47} & 0.61 & \textbf{0.38} & \textbf{0.74} \\
\hline
 T2 & 0.34 & 0.69 & 0.29 & 0.45 \\
\hhline{|=|=|=|=|=|}
 Flair, T1C & 0.47 & 0.81 & 0.47 & 0.78 \\
\hline
 Flair, T1C, T2 & \textbf{0.55} & 0.82 & 0.49 & \textbf{0.8} \\
\hline
 all & 0.49 & \textbf{0.84} & \textbf{0.5} & \textbf{0.8} \\
\hline
\end{tabular}
\end{table}
The concatenation network was used to produce the final results on the BRATS data set, as it was revealed to be the better performing network during experiments on hand MRI. Training was done on five cross-validation folds as described in Section~\ref{chapter:Experiments}. The five networks obtained in this way were then used as an ensemble to segment the BRATS 2015 test set, as well as the BRATS 2013 challenge and leaderboard sets. All of the segmentation maps created by the ensemble were uploaded to an online portal for evaluation. Table~\ref{tab:eval_dice} summarizes the dice scores obtained on these segmentation results, as well as the mean dice score taken over the test sets of all five cross-validation folds (the average test set score of each ensemble member, denoted as ``Cross-validation''). Scores are reported for the regions ``whole'', ``core'' and ``enhanced'', which correspond to the following:
\begin{itemize}
\item Separation of healthy and non-healthy tissue
\item Separation of the tumor core (necrosis, enhanced and non-enhanced tumor) from edema and healthy tissue
\item Separation of the enhanced tumor from the rest of the classes
\end{itemize}
Table~\ref{tab:eval_ppv} and Table~\ref{tab:eval_sens} report the precision and the sensitivity achieved on the same test sets and the same regions. Precision is defined as the probability of a positively classified sample being actually positive, while sensitivity describes the probability of a positive sample being classified as positive. \par

\begin{table}
\centering
\caption{Dice scores obtained on different data sets. \label{tab:eval_dice}}
\begin{tabular}{r c c c}
\multicolumn{1}{c}{ } & \multicolumn{3}{c}{Dice Score} \\
\cline{2-4}
\multicolumn{1}{c}{ } & whole & core & enhanced \\
\cline{2-4}
Cross-validation & 0.9 & 0.79 & 0.44 \\
\hline
BRATS 2015 Test & 0.85 & 0.72 & 0.61 \\
\hline
BRATS 2013 Challenge & 0.87 & 0.78 & 0.71 \\
\hline
BRATS 2013 Leaderboard & 0.78 & 0.64 & 0.57 \\
\end{tabular}
\end{table}

\begin{table}
\centering
\caption{Precision measures obtained on different data sets. \label{tab:eval_ppv}}
\begin{tabular}{r c c c}
\multicolumn{1}{c}{ } & \multicolumn{3}{c}{Precision} \\
\cline{2-4}
\multicolumn{1}{c}{ } & whole & core & enhanced \\
\cline{2-4}
Cross-validation & 0.9 & 0.81 & 0.57 \\
\hline
BRATS 2015 Test & 0.82 & 0.77 & 0.61 \\
\hline
BRATS 2013 Challenge & 0.82 & 0.8 & 0.83 \\
\hline
BRATS 2013 Leaderboard & 0.71 & 0.66 & 0.63 \\
\end{tabular}
\end{table}

\begin{table}
\centering
\caption{Sensitivity measures obtained on different data sets. \label{tab:eval_sens}}
\begin{tabular}{r c c c}
\multicolumn{1}{c}{ } & \multicolumn{3}{c}{Sensitivity} \\
\cline{2-4}
\multicolumn{1}{c}{ } & whole & core & enhanced \\
\cline{2-4}
Cross-validation & 0.92 & 0.87 & 0.68 \\
\hline
BRATS 2015 Test & 0.91 & 0.73 & 0.67 \\
\hline
BRATS 2013 Challenge & 0.95 & 0.8 & 0.67 \\
\hline
BRATS 2013 Leaderboard & 0.93 & 0.73 & 0.59 \\
\end{tabular}
\end{table}
It is clear to see that the performance on the BRATS 2013 Leaderboard set is, in most cases, worse than the performance on the other sets. This is likely due to a difference in difficulty between this set and the rest. With respect to the dice scores obtained on the regions whole and core, the first three sets appear to be in agreement, while the leaderboard results, as mentioned before, are lower. Divergence from this behaviour is observed on the enhanced region, where the lowest score stems from the ensemble average. The precision scores have turned out similarly: whole and core scores of the first three classes appear to agree, while the lowest score for the enhanced region belongs to the cross-validation result. Generally, performance on the enhanced region seems to vary between the four sets, as far as the dice score and the precision are concerned. In comparison to the first two evaluation metrics, sensitivity assumes similar values in all four evaluation sets. \par
\begin{figure}
	\begin{center}
	    \includegraphics[width=0.8\textwidth]{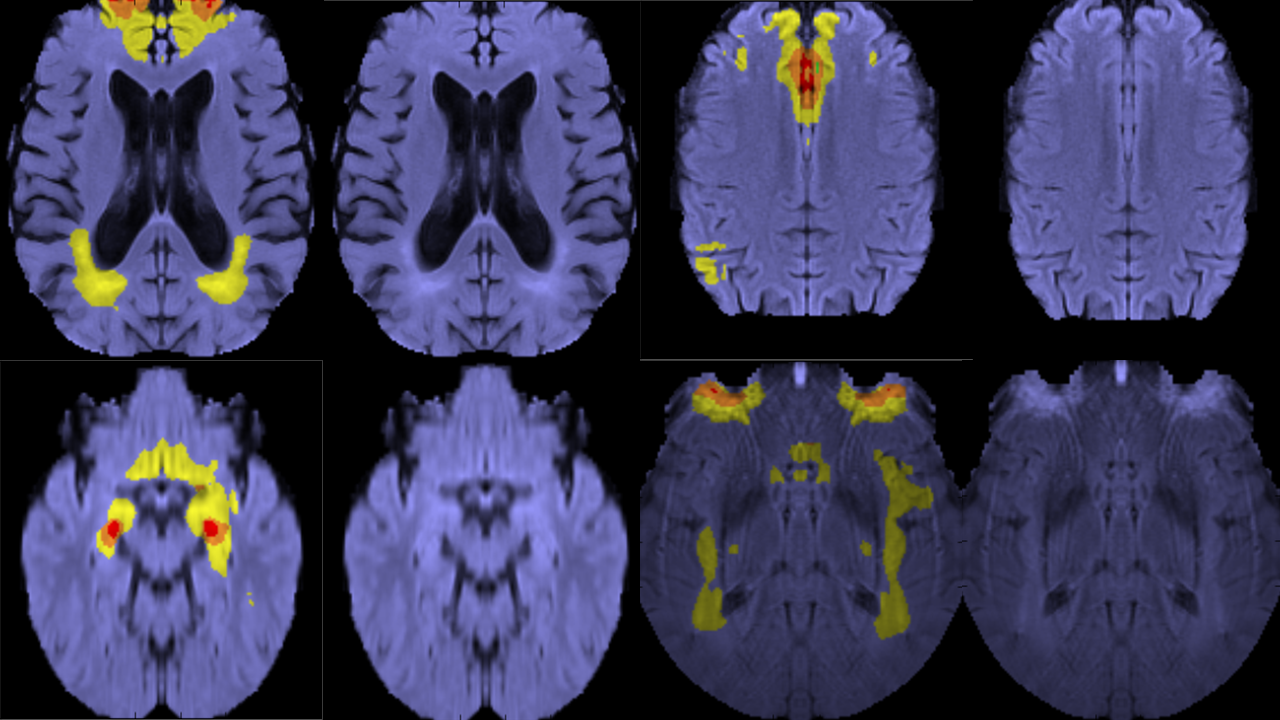}
	\end{center}
	\caption{Instances of false positives on synthetically created healthy images Colors correspond to: necrosis (green), edema (yellow), non-enhanced (red) and enhanced tumor (orange).\label{fig:false_pos}}
\end{figure}
Stable and relatively high values of sensitivity suggest that the behaviour of the model is relatively well-defined for unhealthy tissue. The differences between the leaderboard set and the other three sets on the whole and core regions, as well as the high variance of the dice score and the precision on the enhanced region indicate that the model is prone to false positives. In order to confirm this, the individual ensemble members were tested on synthetic healthy brain images created from the test partition of their respective cross-validation folds. The synthetic brain images are produced by mirroring one hemisphere of the brain that does not include any voxels labelled as unhealthy in the ground truth, for all images where such an operation is possible. These are images where the tumorous tissue is contained within a single hemisphere. \par
The previously used three metrics, dice score, precision and sensitivity, are not suitable in the case of healthy brains, since they largely depend on the existence of true positives (in this case truly unhealthy regions). Specificity, by comparison, is a more suitable metric, as it describes the probability of a healthy voxel being classified as healthy. The mean specificity scores of the five ensemble members are listed in Table~\ref{tab:eval_sh}. \par
Though the mean specificity of the networks appears to be high, visual inspection of the segmentation maps reveals many instances of false positives. These are demonstrated in Fig.~\ref{fig:false_pos}. While some of the false positives can be attributed to artefacts in the input images such as highlighted areas (bottom right image in Fig.~\ref{fig:false_pos}), most suggest that the networks exhibit unexpected behaviour when segmenting images without any visible unhealthy tissue. The model outputs suggest that the networks could benefit from the inclusion of images depicting healthy brains into the data set. This can either be done by creating synthetic images as described before or by acquiring MRI depictions of actual, healthy brains. In the case of this work, the latter could not be done because of the difficulty of finding images in the same MRI modalities as the BRATS data. \par
\begin{figure}
	\begin{center}
	    \includegraphics[width=0.8\textwidth]{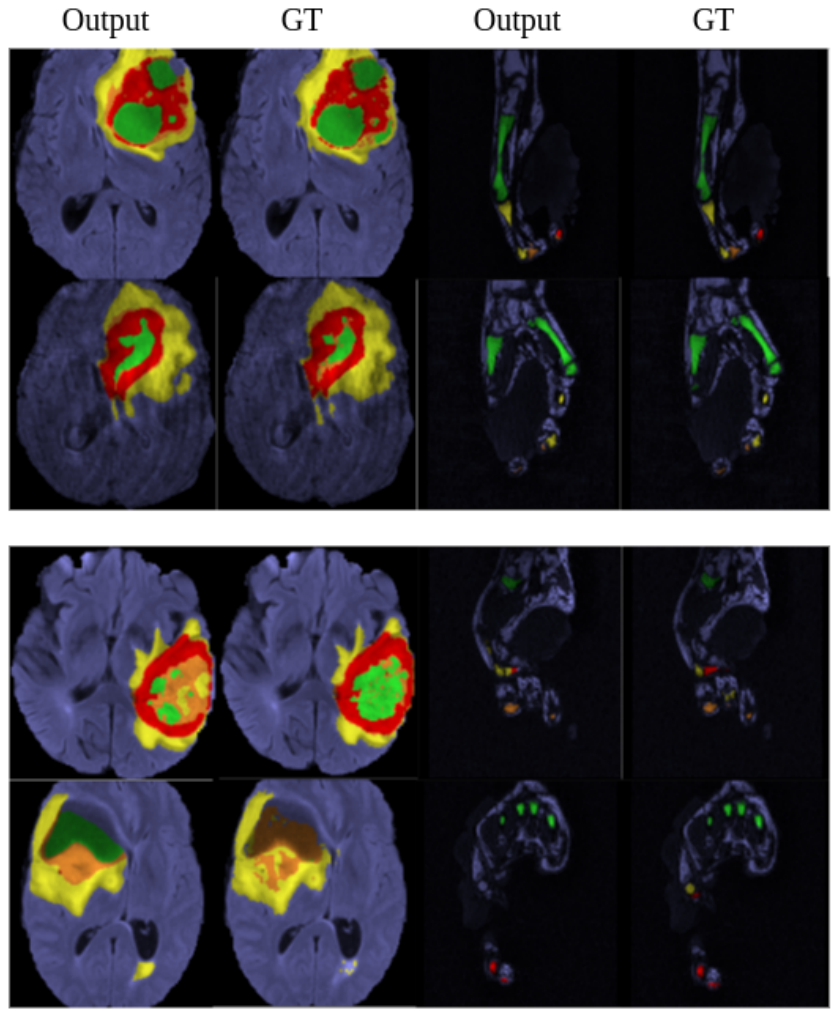}
	\end{center}
	\caption{Model outputs for the hand and brain MRI, depicted alongside the ground truth. The top two rows contain images for which the segmentation has few inaccuracies, while the bottom two rows are representative of the mistakes made by the networks. Colors correspond to: necrosis (green), non-enhanced (red) and enhanced tumor (orange) and edema (yellow) for the brain MRI, and metacarpal (green), proximal (yellow), middle (orange) and distal (red) for the hand MRI \label{fig:output_compilation}}
\end{figure}

\begin{table}
\centering
\caption{Mean specificity of the five ensemble members on synthetic, healthy images. \label{tab:eval_sh}}
\begin{tabular}{r c c c}
\multicolumn{1}{c}{ } & \multicolumn{3}{c}{Specificity} \\
\cline{2-4}
\multicolumn{1}{c}{ } & whole & core & enhanced \\
\cline{2-4}
Cross-validation & 0.9927 & 0.9992 & 0.9993 \\
\end{tabular}
\end{table}
As a conclusion to this section, some model outputs for both the hand and the brain MRI are collected in Fig.~\ref{fig:output_compilation}.

\section{Conclusion}
\label{chapter:Conclusion}
\subsection{Overview of Results}
In this work, a CNN-based medical image segmentation method was demonstrated on hand and brain MRI. Two modifications to the U-Net architecture of Ronneberger et al. \cite{unet} were tested:
\begin{enumerate}
\item Combining multiple segmentation maps created at different scales
\item Using element-wise summation to forward feature maps from one stage of the network to the other
\end{enumerate}
Of these, summation was found to produce slightly worse results, while combining multiple segmentation maps was shown to speed up convergence without hurting the final performance. In order to tackle the issues of class imbalance, which is a challenging aspect of medical image segmentation, a loss function based on the Jaccard similarity index was used. The high memory demand of 3D images were addressed by downsampling the images whenever necessary, as opposed to dividing them into multiple sections, which is another prevalent approach \cite{unet,kamnitsas,voxresnet}. The results demonstrate that, despite the relative scarcity of labelled medical images, 3D CNN architectures are capable of achieving good quality results.

\subsection{Future Work}
The results presented in Subsection~\ref{section:BrainRes} suggest that the networks may benefit from the inclusion of healthy brain MRI into the data set. This will be tested in future research. As was stated before in Subsection~\ref{section:HandRes}, class imbalance can also be addressed by associating weights with the individual classes, which was avoided in this work, as it poses an additional hyper-parameter optimization problem. However, in order to truly justify the usage of alternative loss functions such as the Jaccard Loss, a comparison to weighted categorical cross-entropy needs to be made. Furthermore, it was noted in Section~\ref{section:Data}, that, in the case of the BRATS data set, class imbalance can be made less severe by cropping regions depicting empty space. A comparison between the Jaccard Loss and categorical cross-entropy needs to be made in this case, in order to determine the usefulness of alternative loss metrics in less severely imbalanced scenarios. \par
Another interesting avenue of future research is applying the network to different types of medical images depicting different structures and using different kinds of imaging technology (ultrasound, EM, CT, etc.). Finally, using deeper architectures with more contracting and expanding blocks will be part of future work.

\begin{appendices}
\section{ }
\label{app:A}
Figure~\ref{fig:exact} illustrates the exact network used throughout the experiments. A few details regarding the implementation of batch normalization and deconvolution are necessary. For batch normalization, running averages of the mean and standard deviation of each feature map are computed during training to be used during validation and at inference time \cite{bn}. The running averages are initialized and updated in the following way:
\begin{equation}
\begin{split}
\mu^*_{i+1} & = \alpha * \mu^*_{i} + (1 - \alpha) * \mu_{i}\\
\sigma^*_{i+1} & = \alpha * \sigma^*_{i} + (1 - \alpha) * \sigma_{i}\\
\mu^*_{0} & = 1\\
\sigma^*_{0} & = 0
\end{split}
\end{equation}
Where $\mu^*_i$ and $\sigma^*_i$, as well as $\mu_i$ and $\sigma_i$, denote the running and current metrics respectively. In all of the experiments, $\alpha$ is set to be 0.5. Roughly speaking, this is a compromise between the two extreme strategies of letting new samples have a high influence on the running averages (corresponding to an $\alpha$ close to 0) and the reverse (making it difficult for each new sample to influence the running average, corresponding to an $\alpha$ close to 1). \par
\begin{figure}
	\centering
	\begin{tikzpicture}

\fill[blue!40!white] (1*0.35,0-5*0.35) rectangle (1*0.35+0.35,0-4*0.35);
\fill[blue!40!white] (0*0.35,-3*0.35) rectangle (3*0.35,-2*0.35); 
\fill[blue!40!white] (1*0.35,0) rectangle (3*0.35,0.35);

\fill[red!40!white] (-2*0.35,0-5*0.35) rectangle (-1*0.35,0-4*0.35);
\fill[red!40!white] (-3*0.35,-3*0.35) rectangle (0*0.35,-2*0.35); 
\fill[red!40!white] (0*0.35,0) rectangle (1*0.35,0.35);

\foreach \i in {0,...,3}
{
\draw (\i*0.35,0) rectangle (\i*0.35+0.35,0.35);
}

\draw[->] (0+0.35/2,0) -- (-2*0.35+0.35/2,-2*0.35);
\draw[->] (0.35+0.35/2,0) -- (0+0.35/2,-2*0.35);
\draw[->] (2*0.35+0.35/2,0) -- (2*0.35+0.35/2,-2*0.35);
\draw[->] (3*0.35+0.35/2,0) -- (4*0.35+0.35/2,-2*0.35);

\draw[->] (0+0.35/2,0) -- (-1*0.35+0.35/2,-2*0.35);
\draw[->] (0.35+0.35/2,0) -- (1*0.35+0.35/2,-2*0.35);
\draw[->] (2*0.35+0.35/2,0) -- (3*0.35+0.35/2,-2*0.35);
\draw[->] (3*0.35+0.35/2,0) -- (5*0.35+0.35/2,-2*0.35);

\foreach \i in {-2,...,5}
{
\draw (\i*0.35,0-3*0.35) rectangle (\i*0.35+0.35,0-2*0.35);
}
\draw[dotted] (-3*0.35,0-3*0.35) rectangle (-2*0.35,0-2*0.35) node[pos=.5] {0};
\draw[dotted] (6*0.35,0-3*0.35) rectangle (7*0.35,0-2*0.35) node[pos=.5] {0};

\foreach \i in {-2,...,5}
{
\draw (\i*0.35-0.35,0-3*0.35) -- (\i*0.35,0-3*0.35-0.35);
}
\foreach \i in {-1,...,6}
{
\draw (\i*0.35,0-4*0.35) -- (\i*0.35+0.35,0-4*0.35+0.35);
}

\foreach \i in {-2,...,5}
{
\draw (\i*0.35,0-5*0.35) rectangle (\i*0.35+0.35,0-4*0.35);
}

\end{tikzpicture}
	\caption{1D deconvolution as per the implementation used in this work. Zero-padding is used to prevent pixel loss due to the convolution. Pixels on the corners only ``see'' one pixel of input, while pixels in the middle are influenced by two neighboring pixels each.\label{fig:deconv}}
\end{figure}
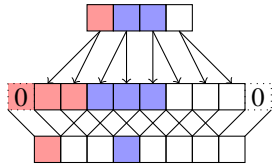
\begin{figure}
	\begin{center}
	    \includegraphics[width=1.\textwidth]{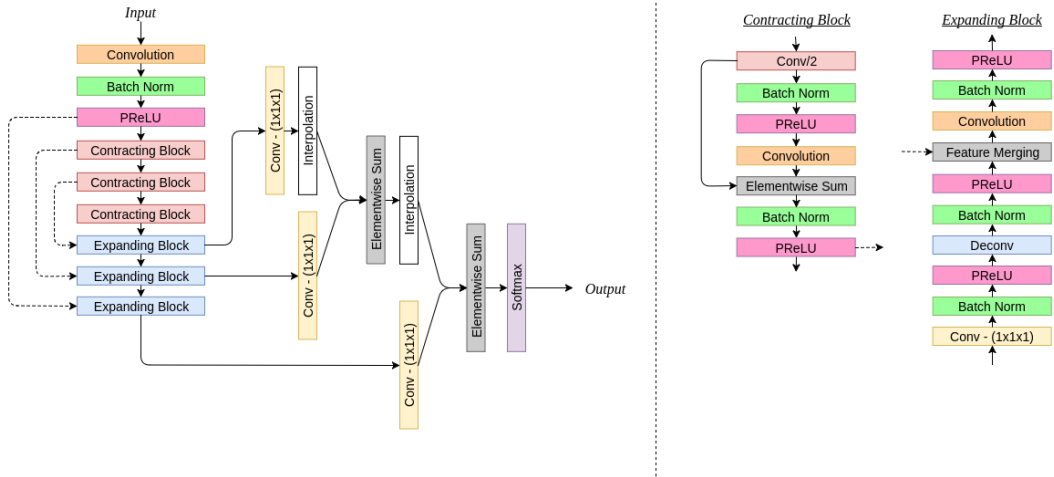}
	\end{center}
	\caption{The CNN used throughout the experiments. All convolutions use $3\times3\times3$ filters, except otherwise stated. The $1\times1\times1$-convolutions at the beginning of each expanding block halve the number of feature maps of their input. The ``Conv$/2$'' layer at the beginning of each contracting block is a 2-strided convolution which halves its input in each dimension. Deconvolution always doubles the size of its input in each dimension (reversing the effect of strided convolution). Feature merging is implemented with element-wise summation or concatenation of the feature maps, depending on the network. \label{fig:exact}}
\end{figure}

The deconvolution operation is implemented by nearest neighbors upsampling followed by convolution. Each voxel is repeated once in every dimension, creating a coarsely upsampled volume, which is then smoothed by convolution. A 1D example of this type of deconvolution is depicted in Fig.~\ref{fig:deconv}. \par
The network is implemented using Theano \cite{theano} and Breze\footnote{The source code of Breze is available at: \url{https://github.com/breze-no-salt/breze}}. The optimization process is done using Climin \cite{climin}.

\section{ }
\label{app:B}
\begin{table}
\centering
\caption{Receptive fields of the convolutional layers present in the network, missing only the convolutions with $1\times1\times1$-sized filters. The input size is chosen for demonstration purposes, actual input volumes used in the experiments vary. \label{tab:rfield}}
\begin{tabular}{|c|c|c|c|c|}
\hline
     Convolution & Input & Output & Receptive Field & Features\\
\hline
     1. & $128\times128\times96$ & $128\times128\times96$ & $3\times3\times3$ 
     & 8\\
\hline
     2. & $128\times128\times96$ & $64\times64\times48$ & $5\times5\times5$ 
     & 8\\
\hline
     3. & $64\times64\times48$ & $64\times64\times48$ & $9\times9\times9$ 
     & 16\\
\hline
     4. & $64\times64\times48$ & $32\times32\times24$ & $13\times13\times13$
     & 32\\
\hline
     5. & $32\times32\times24$ & $32\times32\times24$ & $21\times21\times21$ 
     & 32\\
\hline
     6. & $32\times32\times24$ & $16\times16\times12$ & $29\times29\times29$
     & 64\\
\hline
\multicolumn{5}{|c|}{End of contracting path} \\
\hline
     7. & $16\times16\times12$ & $16\times16\times12$ & $45\times45\times45$
     & 64\\
\hline
\multicolumn{5}{|c|}{Begin of expanding path} \\
\hline
     9. & $16\times16\times12$ & $32\times32\times24$ & $53\times53\times53$ 
     & 32\\
\hline
     10. & $32\times32\times24$ & $32\times32\times24$ & $61\times61\times61$ 
     & 64\\
\hline
     12. & $32\times32\times24$ & $64\times64\times48$ & $65\times65\times65$ & 16\\
\hline
     13. & $64\times64\times48$ & $64\times64\times48$ & $69\times69\times69$ & 32\\
\hline
     15. & $64\times64\times48$ & $128\times128\times96$ & $71\times71\times71$ & 8\\
\hline
     16. & $128\times128\times96$ & $128\times128\times96$ & $73\times73\times73$ & 16\\
\hline
\end{tabular}
\end{table}
As was stated before in the first chapter, a successful CNN-based model for segmentation needs to be able to learn global features of its input. The cumulative receptive field of a CNN is the number of input pixels that determine the value of one output pixel. While a large network receptive field provides no guarantee of increased accuracy \cite{inception}, monitoring the receptive field provides an insight into how much of the global context provided in the input image can theoretically be considered when predicting the class of each pixel. Successive convolutional layers increase the network receptive field. For instance, a pair of convolutional layers with a filter size $3\times3\times3$ have a cumulative receptive field of the size $5\times5\times5$. Down- and upsampling layers change the way that the cumulative receptive field grows with each new convolutional layer. Specifically, the network receptive field at any convolutional layer $i$ in the network depicted in Fig. \ref{fig:network} can be calculated using the following recursive equation:
\begin{equation}
\label{eq:recrfield}
\begin{split}
\varphi_{i} & = \varphi_{i-1} + 2^{(\eta - \tau)} * (\text{Filter size} - 1)\\
\varphi_{0} & = 1
\end{split}
\end{equation}
Where $\eta$ and $\tau$ are the number of strided convolutions and deconvolutional layers used in the network\footnote{In the case of deconvolution, this includes layer $i$ itself.}, respectively. It is assumed that convolutions use zero-padding to preserve the size of their input. It is important to note that Eq.~(\ref{eq:recrfield}) is not a general formulation and only applies to the network used in this work. In general, the way the network receptive field changes with each convolution, deconvolution, and strided convolution depends on the exact implementation of these operations. In the case of this work, convolution stride is always applied to the result of a normal convolution by discarding every $k$-th voxel in every dimension for a stride of $k$. Deconvolution by a factor of $k$ is implemented by repeating every voxel in each dimension $k$ times and following this result by a convolution. Table \ref{tab:rfield} details all the convolutional and deconvolutional layers of the network used in the experiments.
\end{appendices}

\bibliographystyle{habbrv}
\bibliography{bibliography}

\end{document}